\pdfoutput=1

\documentclass[11pt]{article}

\usepackage[final]{acl}

\usepackage{times}
\usepackage{latexsym} 
\usepackage{tikz}
\newcommand*\circled[1]{\tikz[baseline=(char.base)]{
            \node[shape=circle,draw,inner sep=2pt] (char) {#1};}}
\usepackage{booktabs}
\usepackage{graphicx}
\usepackage{multirow}
\usepackage{enumitem}
\usepackage[T1]{fontenc}
\usepackage[utf8]{inputenc}

\usepackage{microtype}

\usepackage{inconsolata}
\title{AILS-NTUA at SemEval-2024 Task 9: Cracking Brain Teasers: Transformer Models for Lateral Thinking Puzzles}

\author{
    Ioannis Panagiotopoulos,  Giorgos Filandrianos, Maria Lymperaiou, Giorgos Stamou\\ 
    School of Electrical and Computer Engineering,  AILS Laboratory\\
    National Technical University of Athens \\
    \texttt{\href{mailto:yiannispn@gmail.com}{yiannispn@gmail.com}, \{\href{mailto:geofila@islab.ntua.gr}{geofila}, \href{mailto:marialymp@islab.ntua.gr}{marialymp}\}@islab.ntua.gr, \href{mailto:gstam@cs.ntua.gr}{gstam@cs.ntua.gr}}\\
}

\begin{document}
\maketitle
\begin{abstract}
In this paper, we outline our submission for the SemEval-2024 Task 9 competition: 'BRAINTEASER: A Novel Task Defying Common Sense'. We engage in both sub-tasks: Sub-task A-Sentence Puzzle and Sub-task B-Word Puzzle. We evaluate a plethora of pre-trained transformer-based language models of different sizes through fine-tuning. Subsequently, we undertake an analysis of their scores and responses to aid future researchers in understanding and utilizing these models effectively. Our top-performing approaches secured competitive positions on the competition leaderboard across both sub-tasks. In the evaluation phase, our best submission attained an average accuracy score of 81.7\% in the Sentence Puzzle, and 85.4\% in the Word Puzzle, significantly outperforming the best neural baseline (ChatGPT) by more than 20\% and 30\% respectively.
\end{abstract}

\section{Introduction}
In Natural Language Processing (NLP), reasoning serves as the cognitive backbone, enabling systems to transcend mere language comprehension and delve into sophisticated understanding.
Despite the excellence of Large Language Models (LLMs) in several linguistic tasks, their reasoning capabilities are still questionable to a non-negligible extend \cite{gpt3-limits, stochastic, unlikely, zhang2023language, distracted, tyen2024llms, giadikiaroglou2024puzzle}, often posing the fundamental concerns of whether they can indeed reason or memorize exhaustively \cite{deductive}. 

Such limitations can be probed via well-crafted datasets and benchmarks, showcasing varying LLM deficiencies at a time. As the core of the current paper, BrainTeaser \cite{jiang2023brainteaser, jiang-semeval-2024-brainteaser} incorporates problems that stress models to think "out-of-the-box"; to this end, the key novelty of BrainTeaser is that in order to answer correctly, models need to defy default senses of concepts and common associations. 
Surprisingly, state-of-the-art (SoTa) LLMs, such as ChatGPT can only exhibit a maximum accuracy of $\sim$60\% when solving BrainTeaser riddles, demonstrating an inherently limited reasoning ability in unconventional thinking.

Thus, assuming that large-scale training and prompting may not always serve as universally applicable solutions
towards flexible reasoning, we move one step back and leverage transfer learning techniques starting from smaller models based on masked language modelling, such as BERT \cite{devlin-etal-2019-bert} and consequent BERT-based encoders. Then, we proceed with similar techniques on LLMs, aiming to showcase that significant performance advancements using a small set of in-domain data for parameter updating can be achieved in comparison to merely querying the model's prior knowledge via prompting. Therefore, our contributions are:
\begin{enumerate}
    \item We perform lightweight tuning on smaller encoder models and LLMs, significantly outperforming the reported baselines.
    \item We transform the multiple-choice problem to a binary classification one, aiming to explore diverging reasoning paths for models.
    \item We ground final performance on the models' "prior knowledge" in related problems.

    \item We delve into models' frequent failures to obtain a deeper understanding of reasoning cues that make models struggle the most.
\end{enumerate}
Our code is available on GitHub \footnote{\url{https://github.com/GiannisPana/AILS-NTUA-at-SemEval-2024-Task-9-Brainteaser}}.


\section{Related work}
\paragraph{Reasoning in NLP} has enjoyed several advancements due to the surge of pre-trained language models and especially LLMs \cite{sun2023survey}. Reasoning challenges incorporate commonsense reasoning \cite{richardson2023commonsense}, involving inference regarding everyday situations, mathematical reasoning \cite{lu-etal-2023-survey}, referring to the ability of solving mathematical problems, logical reasoning \cite{yang2023logical}, which includes the systematic deduction of conclusions based on established principles and formal rules, causal reasoning \cite{gendron2024survey}, which studies cause-and-effect relationships explaining why an event leads to another, and several other sub-tasks \cite{vashishtha-etal-2020-temporal, wei2023enhancing, petersen2023language}. 
In terms of reasoning evaluation, BigBench \cite{srivastava2023imitation} comprises 204 reasoning tasks, targeting to explore the related capabilities of recent LLMs.
Several dedicated datasets have been developed to tackle different reasoning challenges, including commonsenseQA \cite{talmor-etal-2019-commonsenseqa}, WinoGrande \cite{sakaguchi2019winogrande}, RiddleSense \cite{lin2021riddlesense} and others;  most of these datasets are incorporated in Tasksource \cite{sileo2023tasksource}. Especially RiddleSense questions aspects of reasoning close to BrainTeaser \cite{jiang2023brainteaser, jiang-semeval-2024-brainteaser}.

\section{Task and Dataset Description}

The BrainTeaser task at SemEval-2024 \citep{jiang2023brainteaser, jiang-semeval-2024-brainteaser} features lateral thinking puzzles presented as multiple-choice questions (QAs). Each question offers four options, with one being the correct answer and the others serving as distractors. Additionally, the final option is always "None of above". It consists of two sub-tasks, \textit{Task A}: \textbf{Sentence Puzzle} and \textit{Task B}: \textbf{Word Puzzle}. In addition to the original puzzles, the dataset includes adversarial subsets created by manually modifying the original brain teasers while preserving their reasoning paths. The original data were perturbed in two ways: First, there is \textit{semantic reconstruction} of each original question without altering the answers or the distractors. Second, the original data underwent \textit{context reconstruction}, wherein the original reasoning path remains intact, but the brain teaser describes a new situational context. Overall, the dataset used for training and evaluation consists of triplets of data: original, semantic, and context reconstruction. Table \ref{tab:dataset_triplets} provides an example of the triplets of data that constitute the dataset.

 \begin{table}[ht!]
    \resizebox{\columnwidth}{!}{%
    \begin{tabular}{ll}
    \toprule
    \multicolumn{1}{c}{\textbf{Question}} & \multicolumn{1}{c}{\textbf{Choice}} \\
    \hline
    \multicolumn{2}{c}{\textit{Original}} \\
    \hline
     & A peanut. \\
    \cline{2-2} 
    What kind of nut has no shell? & \textbf{A doughnut.} \\
    \cline{2-2} 
     & A walnut. \\
    \cline{2-2} 
     & None of above. \\
    \hline
    \multicolumn{2}{c}{\textit{Semantic Reconstruction}} \\
    \hline
    & \textbf{A doughnut.} \\
    \cline{2-2} 
    Which nut doesn't have a shell? & A walnut. \\
    \cline{2-2} 
     & A peanut. \\
    \cline{2-2} 
     & None of above. \\
    \hline
    \multicolumn{2}{c}{\textit{Context Reconstruction}} \\
    \hline
    & A fire bell. \\
    \cline{2-2} 
    Which type of bell doesn't make a sound? & A cow bell. \\
    \cline{2-2} 
     & \textbf{A bluebell.} \\
    \cline{2-2} 
     & None of above. \\
    \hline
    \end{tabular}%
    }
    \caption{Illustration of the structure of each sub-task's dataset, showcasing the original statement along with its two adversarials.}
    \label{tab:dataset_triplets}
\end{table}

\paragraph{\textit{Task A}: Sentence Puzzle}
In this sub-task, the sentence pairs are crafted in a manner that makes it relatively easy for humans to discern the correct statement, yet challenging for systems, even those equipped with commonsense understanding. Table \ref{tab:example_combined}
contains examples of the Sentence Puzzle dataset (on the left).
The training data consists of 169 distinct multiple-choice QA sets, each accompanied by its semantic and context reconstructions, resulting in a total of 507 multiple-choice questions ($3 \times 169$). 

\paragraph{\textit{Task B}: Word Puzzle} involves word-type brain teasers, where the answer defies the default meaning of the word and focuses on the letter composition of the question. 
The training dataset comprises 132 multiple-choice QAs, each accompanied by its semantic and context reconstructions, resulting in a total of 396 multiple-choice QAs ($3 \times 132$). These brain teaser categories include puns, homophones, ambiguous words, and various other linguistic puzzles, as showcased in the examples provided in Table \ref{tab:example_combined} on the right-hand side.
The Word Puzzle sub-task pose challenges not only for systems but also for humans in discerning the correct answer. 

\begin{table*}[!ht]
    \centering
    \resizebox{\textwidth}{!}{%
    \begin{tabular}{ll|ll}
    \toprule
    \multicolumn{2}{c|}{\textbf{\textit{Sentence Puzzle}}} & \multicolumn{2}{c}{\textbf{\textit{Word Puzzle}}} \\
    \midrule
    \multicolumn{1}{c}{\textbf{Question}} & \multicolumn{1}{c|}{\textbf{Choice}} & \multicolumn{1}{c}{\textbf{Question}} & \multicolumn{1}{c}{\textbf{Choice}} \\
    \midrule
     & \textbf{He is a barber.} & & Cabbages. \\ \cline{2-2} \cline{4-4} 
    A man shaves everyday, yet keeps his beard long. & He wants to maintain his appearance. & What has toes but no feet or legs? & \textbf{Tomatoes.} \\ \cline{2-2} \cline{4-4} 
     & He wants his girlfriend to buy him a razor. &  & Onions. \\ \cline{2-2} \cline{4-4}  
     & None of above. &  & None of above. \\ 
     \midrule
    You go to the doctor because you're sick, & One and a half hours. & & \textbf{Sea-plus.} \\ \cline{2-2} \cline{4-4} 
    and he gives you three medicines to take & \textbf{Two hours.} & What did the little lobster get on its math test? & Very-bad. \\ \cline{2-2} \cline{4-4} 
    every half hour. How long do the drugs & An hour. & & Very-Good. \\ \cline{2-2} \cline{4-4} 
    keep you going? & None of above. & & None of above. \\ 
    \midrule
     & \textbf{Once.} & & The letter T. \\ \cline{2-2} \cline{4-4} 
    How many times can you deduct 10 from 100? & Infinite time. & What's the beginning of an argument? & \textbf{The letter A.} \\ \cline{2-2} \cline{4-4} 
     & Twice. & & The letter U. \\ \cline{2-2} \cline{4-4} 
     & None of above. & & None of above. \\ 
     \bottomrule
    \end{tabular}%
    }
    \caption{Example questions illustrating both sub-tasks, with correct answers highlighted in bold. Examples on the left pertain to \textit{sub-task A: Sentence Puzzle}, while those on the right correspond to \textit{sub-task B: Word Puzzle}.}
    \label{tab:example_combined}
\end{table*}

\paragraph{Data statistics}
The BrainTeaser dataset comprises 3 data splits, namely train, development (used during the practice phase), and the hidden test set, which was used for evaluation. Statisics are provided in Table~\ref{tab:data_stats}. Throughout the evaluation phase, the leaderboard was kept concealed. 

\begin{table}[ht!]
    \centering
    \small
    \begin{tabular}{lccc}
    \hline
    \textbf{Sub-task} & \textbf{Train} & \textbf{Dev} & \textbf{Test} \\
    \hline
    A - Sentence Puzzle & 507 & 120 & 120 \\
    B - Word Puzzle & 396 & 96 & 96 \\
    \hline
    \end{tabular}
    \caption{Data statistics.}
    \label{tab:data_stats}
\end{table}

\paragraph{Evaluation Metrics}

Both sub-tasks are assessed via accuracy metrics to gauge the performance of participating systems in two ways. First, instance-based accuracy evaluates each question individually, considering original questions and their semantic and context adversarials. This metric provides a detailed understanding of a model's proficiency in reasoning through various scenarios. In contrast, group-based accuracy takes a broader perspective, assessing questions and associated adversarials as cohesive groups. Each group consists of three questions, and a model scores 1 only if it correctly solves \textit{all} questions in a group. This approach evaluates the system's holistic performance in navigating through lateral thinking challenges. The combined use of instance-based and group-based accuracy metrics provides comprehensive insights into the capabilities of participating systems in tackling the complexities of both sub-tasks.

\section{Methods}
We focus on tuning language models belonging into two categories.
First, we fine-tune variations of \textit{encoder} models, namely BERT \cite{devlin-etal-2019-bert}, RoBERTa-large \cite{liu2019roberta} and DeBERTaV3-base \cite{he2023debertav3}, to assess the impact of transfer learning using various datasets requiring similar reasoning abilities, apart from BrainTeaser. We study the problem using the provided \textit{multi-choice} setup, but we also transform it into a \textit{binary} classification task. Secondly, the encoders' results are compared with those obtained from \textit{fine-tuned LLMs} using the BrainTeaser dataset. To achieve this, we fine-tune Llama 2 \cite{touvron2023llama}, Phi-2 \cite{gunasekar2023textbooks} and Mistral-7b \cite{jiang2024mixtral}, which have already demonstrated enhanced reasoning abilities. In this regard, we examine the effect of the model size on our task, which has already been reported in the literature to significantly influence the reasoning abilities of the models \cite{touvron2023llama, wei2022emergent}, along with other tuning hyperparameters. Model details are presented in App. \ref{sec:model-selection}.


\subsection{Encoder models}

\paragraph{Pre-training}
First, we evaluate the effects of the pre-training on our task. Thus, we select two variations of each encoder: the \textit{vanilla} one (using the default pre-trained basis and fine-tuned on BrainTeaser data only) and one that has undergone additional pre-training using supplementary commonsense reasoning datasets before fine-tuned on BrainTeaser. 
In the second case, we use the following pre-trained models: \circled{1} BERT-SE: a BERT-base-uncased version pre-trained on the multiple-choice dataset used in SemEval-2020 Task 4b \cite{wang-etal-2020-semeval} \circled{2} RoBERTa-WNGRD:  a RoBERTa-large  version pre-trained on the WinoGrande dataset, and \circled{3} DeBERTaV3-TS: a DeBERTaV3-base model, pre-trained on diverse commonsense reasoning datasets, and fine-tuned with multi-task learning on over 600 tasks from the Tasksource collection. 

\paragraph{Multi-class Classification task}
This strategy involves treating the problem as  multi-class classification: all four provided options are combined with the given question, and consequently these concatenated inputs are fed into the model, which is fine-tuned to select one of the four options as part of a multi-class classification problem. 

\paragraph{Binary Classification task}
Each sample originally consisting of multiple-choice QAs with four available options, underwent the following transformation: each candidate answer (excluding the "None of above" option) was paired with the question receiving the label 0 if the choice was incorrect, or the label 1 for the opposite. In case all the 3 pairings returned 0, it is directly implied that "None of above" is the correct answer.

\subsection{LLMs}
We demonstrate an in-depth examination of fine-tuning SoTa LLMs (Llama 2, Phi-2, and Mistral-7b) in the context of multi-class classification. Note that during inference, the models prompted to provide an \textit{explanation} along with the label. This experimental step, which we have observed to improve the performance of the model, also provides a qualitative identification of flaws in the models' reasoning process. In our experiments, we explore various combinations of LoRA \cite{hu2021lora} $a$ and $r$ hyperparameters, using values of 16, 32, 64, and 128. For the analysis ahead, LLMs are denoted as model\_r\_a, reflecting these hyperparameters. 
Additional technical information, including  prompting details and specifics about QLoRA hyperparameters, is available in App. \ref{sec:experimental}, \ref{sec:loras_hyper}, \ref{sec:prompt-details}.




\begin{table*}[ht!]
    \centering
    \small
    \begin{tabular}{lcccccc}
    \hline
    \textbf{System} & \textbf{Original} & \textbf{Semantic} & \textbf{Context} & \textbf{Ori. + Sem.} & \textbf{Ori. + Sem. + Con.} & \textbf{Overall} \\
    \hline
    \multicolumn{7}{c}{\textbf{Multi-class classification problem }} \\
    \hline
    \textcolor{gray}{Human} & \textcolor{gray}{.907} & \textcolor{gray}{.907} & \textcolor{gray}{.944} & \textcolor{gray}{.907} & \textcolor{gray}{.889} & \textcolor{gray}{.920}\\
    \textcolor{gray}{ChatGPT} & \textcolor{gray}{.608} & \textcolor{gray}{.593} & \textcolor{gray}{.679} & \textcolor{gray}{.507} & \textcolor{gray}{.397} & \textcolor{gray}{.627} \\
    \textcolor{gray}{RoBERTa-L} & \textcolor{gray}{.435} & \textcolor{gray}{.402} & \textcolor{gray}{.464} & \textcolor{gray}{.330} & \textcolor{gray}{.201} & \textcolor{gray}{.434}\\
    \hline
    \textbf{Mistral-7b\_128\_128} & \textbf{.850} & \textbf{.825} & \textbf{.775} & \textbf{.825} & \textbf{.700} & \textbf{.817} \\
    Mistral-7b\_64\_128 & .850 & .825 & .775 & .825 & .700 & .817 \\
    Mistral-7b\_16\_64 & .800 & .800 & .850 & .750 & .725 & .817 \\
    Mixtral-8x7b\_128\_128 & .850 & .825 & .725 & .800 & .700 & .800 \\
    \cmidrule(lr){2-7}
    Llama 2-7b\_64\_128 & .725 & .650 & .700 & .575 & .475 & .692 \\
    Llama 2-13b\_64\_64 & .665 & .614 & .645 & .550 & .400 & .641 \\
    Llama 2-7b\_64\_64 & .625 & .600 & .675 & .550 & .400 & .633 \\
    Llama 2-7b\_64\_32 & .250 & .250 & .425 & .075 & .000 & .308 \\
    \cmidrule(lr){2-7}
    Phi-2\_64\_128 & .625 & .575 & .550 & .525 & .425 & .583 \\
    Phi-2\_128\_128 & .625 & .575 & .550 & .500 & .375 & .583 \\
    Phi-2\_64\_64 & .525 & .425 & .550 & .375 & .300 & .500 \\
    \cmidrule(lr){2-7}
    RoBERTa-WNGRD & .800 & .775 & .775 & .750 & .675 & .784 \\

    DeBERTaV3-TS & .800 & .775 & .725 & .750 & .625 & .767 \\
    DeBERTaV3-base & .725 & .750 & .675 & .725 & .625 & .717 \\
    BERT-SE & .750 & .725 & .650 & .700 & .550 & .708 \\
    RoBERTa-large & .700 & .700 & .725 & .675 & .550 & .708 \\
    BERT & .675 & .650 & .650 & .600 & .475 & .658 \\
    \hline
    \multicolumn{7}{c}{\textbf{Binary classification problem }} \\
    \hline

    DeBERTaV3-TS & .725 & .650 & .550 & .650 & .650 & .642 \\
    RoBERTa-WNGRD & .575 & .600 & .500 & .550 & .550 & .558 \\
    BERT-SE & .625 & .550 & .375 & .525 & .525 & .517 \\
    \hline
    \end{tabular}
\caption{Model Performance for \textit{sub-task A: Sentence Puzzle}. More results in Table \ref{tab:sentence_results-lora}.}
\label{tab:sentence_results}
\end{table*}

\section{Experimental Results}
Our metrics for the Sentence Puzzle sub-task are presented in Table~\ref{tab:sentence_results} and for the Word Puzzle sub-task in Table~\ref{tab:word_results} along with their baselines.
Interestingly, the performance of the binary classification problem is significantly lower than that of the multi-class classification task. Initially, this behavior seemed counterintuitive since it appeared easier to determine whether a question is correct or not than to select the correct answer from four different options. However, this assumption is not accurate. Consider the word riddle: \textit{`What is the capital in France?"} At first glance, the option `F' seems incorrect, but when considering the options `F,' `E', `A', and `None of the above', `F' emerges as the only correct answer, as it becomes apparent that the question refers to the capital \textit{letter} rather than the capital \textit{city}. Therefore, the diverse options provide crucial context to the models, explaining the superior performance of multi-class models. This lack of context is why we refrain from further exploring this methodology across all models in our study.

\paragraph{\textit{Task A}: Sentence Puzzle}
Table \ref{tab:sentence_results} illustrates minimal fluctuations among all instance-based metrics. This consistency extends to the associated group-based metrics for all models, highlighting a systematic behavior towards detecting various reasoning paths.
This observation holds for both the encoder-based classifiers and LLMs utilized in this sub-task.
Sentence puzzles inherently offer more detailed information, enabling models to detect and identify the same reasoning patterns more readily, regardless of changes in context, in contrast to word puzzles, which typically feature shorter contextual statements, presenting a greater challenge for models to discern consistent reasoning patterns.

Initially, it becomes apparent that pre-training encoders across various commonsense reasoning datasets results in substantial performance enhancements, as it enables the system to grasp domain-agnostic features which prove advantageous for the subsequent task. Additionally, several commonsense pre-trained encoders fine-tuned on BrainTeaser data outperform Llama 2 and Phi-2.

Another noteworthy observation from Table \ref{tab:sentence_results} is that only Mistral-7b from LLMs is able to surpass the encoder-type networks, while both Llama 2 and Phi-2 consistently scored lower. 
Unlike Llama 2 and Mistral-7b, Phi-2 has not undergone instruction fine-tuning \cite{gunasekar2023textbooks}, which, coupled with the limited number of examples in the  BrainTeaser Sentence Puzzle dataset, contributes to its lower performance, as a result of Phi's incapability to capture the complexities of the BrainTeaser data.
In this regard, Mistral-7b, which has already demonstrated superior performance compared to every Llama 2 variation when tested in commonsense reasoning benchmarks \cite{jiang2023mistral}, is also capable of solving this task more accurately.


\begin{table*}[ht!]
    \centering
    \small
    \begin{tabular}{lcccccc}
    \hline
    \textbf{System} & \textbf{Original} & \textbf{Semantic} & \textbf{Context} & \textbf{Ori.+Sem.} & \textbf{Ori.+Sem.+Con.} & \textbf{Overall} \\
    \hline
    \multicolumn{7}{c}{\textbf{Multi-class classification problem }} \\
    \hline

    \textcolor{gray}{Human} & \textcolor{gray}{.917} & \textcolor{gray}{.917} & \textcolor{gray}{.917} & \textcolor{gray}{.917} & \textcolor{gray}{.900} & \textcolor{gray}{.917} \\
    \textcolor{gray}{ChatGPT} & \textcolor{gray}{.561} & \textcolor{gray}{.524} & \textcolor{gray}{.518} & \textcolor{gray}{.439} & \textcolor{gray}{.292} & \textcolor{gray}{.535} \\
    \textcolor{gray}{RoBERTa-L} &  \textcolor{gray}{.195} & \textcolor{gray}{.195} & \textcolor{gray}{.232} & \textcolor{gray}{.146} & \textcolor{gray}{.061} & \textcolor{gray}{.207} \\
    \hline
    \textbf{Mistral-7b\_16\_64} & \textbf{.875} & \textbf{.906} & \textbf{.781} & \textbf{.813} & \textbf{.719} & \textbf{.854}\\
    Mistral-7b\_128\_128 & .844 & .844 & .813 & .719 & .625 & .833\\
    Mistral-7b\_8\_16 & .781 & .938 & .781 & .719 & .562 & .833 \\   
    Mixtral-8x7b\_128\_128 & .625 & .719 & .625 & .531 & .375 & .656 \\
    \cmidrule(lr){2-7}
    Llama 2-13b\_64\_64 & .354 & .344 & .438 & .125 & .031 & .379 \\
    Llama 2-7b\_64\_64 & .375 & .344 & .375 & .125 & .031 & .365 \\
    Llama 2-7b\_64\_128 & .281 & .188 & .438 & .031 & .031 & .302 \\
    \cmidrule(lr){2-7}
    Phi-2\_64\_64 & .688 & .625 & .688 & .562 & .438 & .667 \\
    Phi-2\_64\_128 & .656 & .656 & .625 & .594 & .406 & .646 \\
    Phi-2\_16\_64 & .625 & .500 & .688 & .438 & .312 & .604 \\
    \cmidrule(lr){2-7}
    DeBERTaV3-base & .750 & .750 & .562 & .656 & .438 & .687 \\
    DeBERTaV3-TS & .812 & .781 & .406 & .719 & .281 & .666 \\
    RoBERTa-WNGRD & .750 & .656 & .500 & .625 & .312 & .635 \\
    BERT & .562 & .594 & .469 & .562 & .312 & .542 \\
    BERT-SE & .562 & .500 & .406 & .500 & .281 &  .489\\
    RoBERTa-large & .281 & .344 & .344 & .156 & .094 & .323 \\
    \hline
    \multicolumn{7}{c}{\textbf{Binary classification problem }} \\
    \hline
    DeBERTaV3-TS & .250 & .313 & .313 & .219 & .219 & .292 \\
    RoBERTa-WNGRD & .375 & .313 & .125 & .281 & .281 & .271 \\
    BERT-SE & .000 & .000 & .000 & .000 & .000 & .000 \\

    \hline
    \end{tabular}
\caption{Model performance for \textit{sub-task B: Word Puzzle}. More results in Table \ref{tab:sentence_results-lora}.}
\label{tab:word_results}
\end{table*}
\paragraph{\textit{Task B}: Word Puzzle}
In Table~\ref{tab:word_results}, we observe a stark contrast in the models' performance in understanding and detecting reasoning paths when the context changes. 
There are notable discrepancies in accuracy between original and semantic contexts when compared to context reconstruction, particularly evident in the case of smaller encoder models.
\begin{table*}[ht!]
    \centering
    \small
    \resizebox{\linewidth}{!}{%
    
    \begin{tabular}{p{0.35\linewidth} p{0.55\linewidth}}
    \toprule
    \multicolumn{2}{c}{\textit{Example of correct prediction with explanation}} \\
    \hline
    \multicolumn{1}{c}{\textbf{Question}} & \multicolumn{1}{c}{\textbf{Choice}} \\ 
    \midrule
     & \textbf{The other end is not tied to anything.} \\ 
    \cline{2-2} 
    If a dog is tied to a piece of rope that is 6m long how can it reach a bone that is 7m away?  & The rope stretches proportionally, providing the extra length needed for the horse to reach the hay seven meters away. \\ 
   \cline{2-2} 
     & The dog dug a 1m tunnel underground to reach the bone that was 7m away while still tied to the 6m long rope. \\ 
    \cline{2-2} 
     & None of above. \\ 
    \midrule
    \multicolumn{2}{p{0.9\textwidth}}{\textbf{Explanation:}  A dog cannot reach a bone that is 7m away if it is tied to a piece of rope that is 6m long. This is because the distance between the dog and the bone is greater than the length of the rope. Therefore, the dog will not be able to reach the bone unless the rope is untied.}\\
    \midrule
    \multicolumn{2}{c}{\textit{Example of misprediction with explanation}} \\
    \hline
    \multicolumn{1}{c}{\textbf{Question}} & \multicolumn{1}{c}{\textbf{Choice}} \\ 
    \midrule
     & You need to drop the egg gently. \\ 
    \cline{2-2} 
    How can you drop a raw egg onto a concrete & You need to drop the egg hard. \\ 
    \cline{2-2} 
    floor and not crack it?   & \textbf{No problem, because concrete floors are very hard to crack.} \\ 
    \cline{2-2} 
      & None of above. \\ 
    \midrule
    \multicolumn{2}{p{0.9\textwidth}}{\textbf{Predicted Answer:}  None of above.}\\
    \multicolumn{2}{p{0.9\textwidth}}{\textbf{Explanation:}  There is no way to drop a raw egg onto a concrete floor and not crack it. Concrete is a very hard material and will easily break any object that falls on it. The only way to avoid this would be to use some sort of protective covering for the egg before dropping it.}\\
    \midrule
    \end{tabular}%
    }
    \caption{An example of explanations generated by Mistral-7b, while addressing the multi-class classification problem. The explanations, regardless of their validity, are detailed and showcase a high level of comprehension and reasoning. Correct answers are indicated in bold, while mispredictions reference the predicted answer by the model.}
    \label{tab:explanations}
\end{table*}

Regarding encoders, it is evident that, especially vanilla RoBERTa-large lacks robust commonsense reasoning and struggles to systematically handle ambiguity; in contrast, RoBERTa-large pre-trained on WinoGrande presents competitive performance.
This notable enhancement (over 40\%) due to WinoGrande pre-training suggests that this particular dataset effectively equips the model with the ability to understand word puzzle-related reasoning complexities,  making its scores competitive with DeBERTaV3 in this sub-task, despite the higher DeBERTaV3-base performance over RoBERTa-large in baseline reasoning benchmarks \cite{he2023debertav3}.
Other than that, pre-training on other commonsense reasoning datasets does not significantly improve the overall performance for encoders. Conclusively, apart from WinoGrande the rest of the extra pre-training datasets do not hold reasoning cues close to BrainTeaser's word puzzles.

Regarding LLMs, Mistral-7b notably outperformed all others by a significant margin, even surpassing the 8 times larger model tuned using the same hyperparameters (\textit{Mixtral-8x7b}
). Llama 2 exhibited the worst results regardless of size (7/13 billion) and LoRA hyperparameters (\textit{r} and \textit{a}). Conversely, Phi-2 demonstrated relatively better performance, particularly considering its smaller parameter count (2.7 billion) compared to the other LLMs. However, both models performed worse compared to most fine-tuned encoders. This observations strongly confirms that word puzzles possess a distribution that diverges from the analytical commonsense reasoning required for sentence puzzles, entailing a unique set of cognitive demands.

Mistral-7b exhibits a trend where higher quality explanations were generated with higher values of lora rank \textit{r}. 
However, the top-performing model showcased a configuration with \textit{r}=16 and \textit{a}=64. The QLoRA method \cite{hu2021lora} explains why our top model has a rank of 16 instead of 128, contrary to common expectations (more details reagrding QLoRA hyperparameters in App. \ref{sec:loras_hyper}). Drawing from the widespread presence of low-rank structures, as highlighted by prior studies \cite{li2016recovery, li2019algorithmic,  grasedyck2013literature}, we leverage the intrinsic low-rank structure in our problem, as emphasized in \citet{hu2021lora}. It is well-established that many tasks, particularly involving heavily over-parametrized models, exhibit low-rank properties post-training \cite{oymak2019generalization}. 

Overall, 
our systems demonstrate remarkably high overall accuracy, being less than 10\% lower than human performance and more than 30\% greater than ChatGPT. This suggests our methods' proficiency in understanding and detecting wordplay patterns, consistently addressing ambiguity irrespective of contextual and semantic variations in brain teasers. Upon reviewing the short explanations provided with each prediction (Table~\ref{tab:explanations}), we note thorough justifications even for incorrect answers. Errors typically adhere to specific wordplay patterns across original, semantic, and context multiple-choice questions (details in App. \ref{sec:quality}).

\section{Conclusion}
In this study, we systematically evaluate pre-trained  and fine-tuned encoders, along with instruction-tuned Large Language Models (LLMs), against two multi-class classification sub-tasks within the "BRAINTEASER: A Novel Task Defying Common Sense". We achieve competitive performance in both sub-tasks, accompanied by a plethora of insights regarding the influence of leveraging in-domain data, the variability model scale and architecture introduce, as well as the examination of diverging reasoning paths. As future work, we will delve into further reasoning patterns LLMs tend to follow with regard to lateral thinking challenges.

\bibliography{acl_latex}

\appendix

\section{Model Selection}
\label{sec:model-selection}

\subsection{Encoder}
BERT \cite{devlin-etal-2019-bert}: Bidirectional Encoder Representations for Transformers, is a pretrained deep bidirectional transformer model producing context representations. Using a fine-tuning setting, BERT has advanced state-of-the-art performances on a wide range of NLP tasks.


RoBERTa-large \cite{liu2019roberta}: Robustly Optimized BERT pre-training Approach (RoBERTa) is an adaptation of BERT architecture trained with larger batches on 160 GB data from various domains. RoBERTa-large was trained by dynamically modifying language masking while the next sentence prediction loss used in BERT was
dropped. Other improvising techniques like larger input text sequences, byte pair encoding are used in training which seemingly improved the model performance in downstream tasks. 

DeBERTaV3 \cite{he2023debertav3}: Decoding-enhanced BERT with disentangled attention is an extension of the original DeBERTa model. It builds upon the BERT (Bidirectional Encoder Representations from Transformers) architecture, aiming to enhance its decoding capabilities and overall performance across various natural language processing (NLP) tasks. DeBERTaV3 further improves the efficiency of DeBERTa \cite{he2021deberta} using ELECTRA-Style pre-training with Gradient Disentangled Embedding Sharing. Compared to DeBERTa, V3 significantly improves the model performance on downstream tasks. It incorporates a disentangled attention mechanism to allow the model to focus on different aspects of input independently, improving its ability to capture diverse linguistic patterns. The model also features enhancements in the decoding process, enabling more accurate text generation and sequence classification. 

\subsection{LLMs}
Mistral-7b \cite{jiang2023mistral}: Developed by EleutherAI, is a language model tailored for large-scale natural language processing tasks. With its 7 billion parameters, it excels in handling complex language understanding and generation tasks. Designed to perform exceptionally well across various NLP applications such as text generation, comprehension, and summarization, Mistral-7b surpasses the best open 13b model, Llama 2 \cite{touvron2023llama}, and the best released 34b model, Llama 1  \cite{touvron2023llama1}, in reasoning, mathematics, and code generation tasks. Leveraging grouped-query attention (GQA) and sliding window attention (SWA), Mistral-7b ensures efficient inference and can handle sequences of arbitrary length with reduced inference cost. Its performance across a wide range of benchmarks makes it a promising solution for our sub-tasks, given its extensive task capabilities and superior performance in baseline benchmarks compared to similar or larger language models. While we considered experimenting with its larger variant, Mixtral-8x7b \cite{jiang2024mixtral}, limitations on available resources forced us to deal in depth only with the small variant, Mistral-7b.

Llama 2 \cite{touvron2023llama}: A language model that represents a significant advancement in natural language processing. It is a collection of pre-trained and fine-tuned large language models (LLMs) ranging in scale from 7 billion to 70 billion parameters. With its large parameter count and advanced architecture, Llama 2 is designed to tackle complex language understanding and generation tasks effectively. It outperforms many other models, including its predecessor, Llama 1, in various benchmarks, demonstrating superior capabilities in reasoning, mathematics, and code generation. Leveraging its extensive parameterization and innovative techniques, Llama 2 offers state-of-the-art performance across a wide range of NLP applications, making it a notable contender in the field. For our experiments we were able to experiment with various configurations wit the 7 billion and the 13 billion models. Our involvement with the 70 billion parameter model has been restricted due to limitations associated with the extensive parameter count, particularly during the fine-tuning process.

Phi-2 \cite{gunasekar2023textbooks}: An advanced language model designed to address complex natural language processing tasks efficiently. It is part of the small language models (SLMs) released by Microsoft Research team. With its innovative architecture and extensive parameter count, Phi-2 surpasses its predecessor, Phi-1, in various benchmarks, showcasing superior performance in reasoning, comprehension, and text generation. Leveraging cutting-edge techniques and a comprehensive understanding of language patterns, Phi-2 demonstrates remarkable capabilities across a diverse range of NLP applications, solidifying its position as a prominent model in the field. Given its 2.7 billion-parameter architecture, which exhibits exceptional reasoning and language understanding abilities in comparison to various Llama 2 iterations and Mistral-7b, we are confident that this model will deliver noticeable performance for both of our sub-tasks.

\section{Experimental Setup}
\label{sec:experimental}
In our experiments, we employed the Google Colab platform and Kaggle, leveraging various open-source Python packages such as Transformers, TRL (Transformer Reinforcement Learning) \cite{vonwerra2022trl}, PEFT (Parameter-Efficient Fine-Tuning) \cite{peft}, BitsAndBytes, Accelerate \cite{accelerate}, and Sentence-Transformers.

\paragraph{Encoders}
BERT-SE\footnote{\href{https://huggingface.co/JazibEijaz/bert-base-uncased-finetuned-semeval2020-task4b-append-e3-b32-l4e5}{https://huggingface.co/JazibEijaz/bert-base-uncased-finetuned-semeval2020-task4b-append-e3-b32-l4e5}}: During fine-tuning, a learning rate of 3e$^{-5}$ was used, with a batch size of 16 samples processed in each iteration, over the course of 3 epochs. This process aimed to adapt the pre-trained model to better suit our sub-task. Our optimizer was AdamW and our learning scheduler was linear. Same setup was used for the fine-tuning of the BERT encoder.

RoBERTa-WNGRD\footnote{\url{https://huggingface.co/DeepPavlov/roberta-large-winogrande}} underwent fine-tuning on the train split of each dataset, utilizing a learning rate of 3e$^{-5}$, a batch size of 16, and running for 3 epochs. The opptimizer was also AdamW and the learning scheduler was linear. RoBERTa-large was fine-tuned on the train split of each sub-task's specific dataset using identical configurations.

DeBERTaV3-TS\footnote{\url{https://huggingface.co/sileod/deberta-v3-large-tasksource-nli}}, like DeBERTaV3-base, underwent a fine-tuning process similar to the RoBERTa-WNGRD system, differing only in the batch size, which was set to 4.

\paragraph{LLMs} 
Phi-2\footnote{\url{https://huggingface.co/microsoft/phi-2}} underwent fine-tuning using the prompt format outlined in Section \textit{Prompting Details}. The fine-tuning process involved setting a learning rate of 2e$^{-5}$ and a batch size of 2, with the model trained for 250 steps. We conducted experiments with different configurations of \textit{r} and lora\_alpha, encompassing combinations such as \textit{r} = 64, 128 and lora\_alpha = 64, 128. The dropout rate was consistently set to 0.1 across all experiments. We used an AdamW optimizer and a constant learning scheduler. Despite promising benchmarks accompanying its release, the model's performance during inference on the test split of both sub-tasks' datasets was subpar, scoring lower compared to the encoders mentioned above. This discrepancy raises the possibility, supported by various reports, that the model's training process using methods like quantization and LoRA may not be fully optimized yet, particularly given its recent introduction.

Both variations of Llama 2\footnote{\url{https://huggingface.co/docs/transformers/en/model_doc/llama2}}, with 7 billion and 13 billion parameters, underwent the same fine-tuning pipeline described earlier, utilizing the QLoRA technique. The fine-tuning process followed the prompt format outlined in Section \ref{sec:prompt-details} (\textit{Prompting Details}), employing a learning rate of 2e$^{-5}$ and a batch size of 1, with each model trained for 250 steps. Despite experimenting with various combinations of values for \textit{r} and \textit{a} (32, 64, 128), while the dropout rate was consistently set to 0.1, the results were disappointing. As a text generation model, Llama 2 provided explanations for each multiple-choice prompt. However, even when incorrectly predicting a choice as correct, the generated explanations often lacked logical coherence. Many explanations produced during the inference phase were irrelevant to the context of the brain teaser, indicating a failure to capture the reasoning path of most multiple-choice questions. In summary, both variations of Llama 2, despite their large scale, proved incapable of effectively understanding and reasoning through the multiple-choice questions provided.

The Mistral-7b\footnote{\url{https://huggingface.co/mistralai/Mistral-7B-v0.1}} model outperformed all others significantly. Prior to fine-tuning, we applied the QLoRA technique. Using a learning rate of 2e$^{-5}$ and a batch size of 2, each model underwent fine-tuning for 250 steps using the train split of the sub-tasks' dataset. The initial results were promising.
During experimentation with the \textit{r} and \textit{a} parameters, while maintaining a dropout of 0.1, certain patterns emerged. Specifically, we observed higher quality explanations and scores when using higher rank values, ranging from (16, 32, 64, 128). This outcome was expected, as higher rank values correspond to higher precision weight changes, resulting in superior weight tuning and overall model performance. 
Interestingly, when the ratio of $\textit{a} / \textit{r}$ was low (0.5 - 1), explanations maintained high quality irrespective of predictions, implying a coherent reasoning path even if the predicted choice was incorrect. However, setting the $\textit{a} / \textit{r}$ ratio to 2 or 4 potentially enhanced results, signifying a stronger influence from QLoRA layers on the base model. However, this adjustment led to a decline in the quality of explanations. The improvement could be attributed to the model's low intrinsic dimensionality. Despite having many parameters, the effective dimensionality of the model's learned representations is low. Consequently, after conducting several experiments, the best-performing model regarding word puzzles aligns with this concept.
After conducting numerous tests, we achieved our best performances with the first model using r=128 and alpha=128, and the third best using r=64 and alpha=32. These models are denoted as Mistral-7b\_lora\_r\_lora\_a, representing Mistral-7b\_128\_128 and Mistral-7b\_64\_32 configurations, respectively.

Our exploration of Mistral-8x7b\footnote{\url{https://huggingface.co/mistralai/Mixtral-8x7B-v0.1}} was constrained, yet initial results were promising, despite the limited configurations. Further experimentation with various hyperparameter settings may yield improved performance. In our single attempt with this system, we employed a learning rate of 2e$^{-5}$ and a batch size of 2, fine-tuning the models for 250 steps using the train split of the sub-task's dataset. Both \textit{r} and \textit{a} were set to 128, accompanied by a dropout rate of 0.1. This configuration was selected based on the \textit{r} and \textit{a} values of the best-performing model across both sub-tasks, Mistral-7b. Despite its larger scale, Mistral-8x7b achieved the second-best accuracy during inference on the test split regarding the first subtask, trailing behind its smaller variation, Mistral-7b. This model is referenced in the results table of both sub-tasks as Mistral-8x7b\_128\_128. Further experimentation with various configurations may yield improvements, particularly when leveraging the low intrinsic dimensionality and redundancy inherent in the model.

\section{QLoRA hyperparameters}
\label{sec:loras_hyper}
Initially, we employed the QLoRA technique \cite{dettmers2023qlora} for optimization. The QLoRA technique entails the following steps. First we quantized the models using 4-bit precision to reduce memory usage and computational requirements. The quantization process was facilitated by the BitsAndBytes library. Following quantization, we implemented the LoRA technique \cite{hu2021lora} using the PEFT library. LoRA, applied to the quantized model, resulted in the creation of Quantized LoRA (QLoRA). This pipeline effectively addresses the challenges posed by memory-intensive models on hardware with limited capabilities, ensuring optimized performance and resource utilization.
Regarding the hyperparameters of the QLoRA, the rank \textit{(r)} determines the dimensionality of the low-rank approximation used in the adapter layers, while alpha \textit{(a)} is the scaling factor that determines the magnitude of the newly learned weights compared to the original model's weights. The choice of alpha influences how much emphasis is given to the task-specific information compared to the pre-trained knowledge encoded in the original model. 

In our experiments, we observed that lower values of r occasionally yielded slightly superior results. This phenomenon can be attributed to the regularization effect introduced by lower-rank approximations. Essentially, lower-rank approximations act as a form of regularization, discouraging the model from memorizing the training data and instead promoting the learning of more generalizable patterns. This regularization effect becomes particularly significant when dealing with small datasets, as the risk of overfitting is heightened in such scenarios. By limiting the model's capacity through lower-rank approximations, we encourage it to focus on learning essential features and avoid capturing noise or idiosyncrasies present in the training data. Therefore, in our case where the dataset size is small, the regularization provided by lower-rank approximations becomes crucial. It helps prevent overfitting and encourages the model to generalize better to unseen data, ultimately leading to improved performance in certain cases.

Table~\ref{tab:sentence_results-lora} depicts further analysis of LoRA hyperparameters for Mistral and Mixtral models, which have exhibited the best results among all other models and across the two tasks. Due to computational restrictions, we trained the Mixtral model, which is eight times larger, only for the best performing hyperparameters of Mistral, as a proxy for the performance difference. 

\begin{table*}[ht!]
    \centering
    \small
    \begin{tabular}{lcccccc}
    \hline
    \textbf{System} & \textbf{Original} & \textbf{Semantic} & \textbf{Context} & \textbf{Ori. + Sem.} & \textbf{Ori. + Sem. + Con.} & \textbf{Overall} \\
    \hline
    \multicolumn{7}{c}{\textbf{Task A }} \\
    \hline
    Mistral-7b\_64\_128 & \textbf{.850} & \textbf{.825} & \textbf{.775} & \textbf{.825} & \textbf{.700} & \textbf{.817} \\
    Mistral-7b\_16\_64 & .800 & .800 & .850 & .750 & .725 & .817 \\
    Mixtral-8x7b\_128\_128 & .850 & .825 & .725 & .800 & .700 & .800 \\
    Mistral-7b\_128\_64 & .850 & .800 & .725 & .775 & .625 & .792 \\
    Mistral-7b\_64\_32 & .850 & .775 & .725 & .750 & .675 & .783 \\
    Mistral-7b\_8\_16 & .800 & .800 & .700 & .750 & .625 & .767 \\

    Mistral-7b\_128\_32 & .825 & .775 & .725 & .750 & .600 & .775 \\
    \hline 
    \multicolumn{7}{c}{\textbf{Task B }} \\
    \hline
     Mistral-7b\_128\_128 & \textbf{.844} & \textbf{.844} & \textbf{.813} & \textbf{.719} & \textbf{.625} & \textbf{.833}\\
    Mistral-7b\_8\_16 & .781 & .938 & .781 & .719 & .562 & .833 \\
    Mistral-7b\_16\_16 & .812 & .812 & .875 & .688 & .625 & .833 \\
    Mistral-7b\_8\_8 & .875 & .812 & .812 & .750 & .688 & .833 \\
    Mistral-7b\_16\_32 & .875 & .812 & .781 & .750 & .594 & .823 \\
    Mistral-7b\_64\_32 & .844 &	.875 &	.719 &	.750 &	.562 &	.812\\
    Mistral-7b\_128\_64 & .844 & .812 & .781 & .688 & .531 & .812 \\   
    Mistral-7b\_64\_64 & .719 & .812 & .625 & .625 & .406 & .719 \\
    Mixtral-8x7b\_128\_128 & .625 & .719 & .625 & .531 & .375 & .656 \\
    \hline
    \end{tabular}
\caption{The performance of various LoRA hyperparameters for Mistral and Mixtral in both sub-tasks. }
\label{tab:sentence_results-lora}
\end{table*}

\section{Prompting Details}
\label{sec:prompt-details}
Here, we provide a comprehensive overview of the prompt utilized consistently throughout the fine-tuning process of the LLMs, which ultimately led to optimal performance across both sub-tasks.
Prompt:
\begin{verbatim}
### Instructions:
Below is an instruction that describes a 
multiple choice task. Answer the following 
multiple choice question by giving the 
most appropriate response. Answer should 
be one among options provided after the 
question. Select the most suitable answer 
while making the necessary assumptions. 
Give only answer and a short explanation 
of two or three sentences. Nothing else.

### Input:
Question: {question}
1) {a}
2) {b}
3) {c}
4) {d}

### Answer:
The correct answer is: {label}) {answer}
\end{verbatim}
In the \textit{Instructions} section, we define the task and provide detailed steps for the system. Results varied depending on the content of the \textit{Instructions} section. It's important to note that our model isn't just tasked with selecting the most appropriate choice from the given options; it's also instructed to generate a brief explanation. This additional step aims to assess the model's ability to identify and comprehend a logical reasoning path that can justify its chosen answers for each multiple-choice problem. Given that the questions are brain teasers that challenge common sense, this approach helps us gauge the model's understanding and reasoning capabilities more effectively. In the \textit{Input} section, we structure the provided dataset into a multiple-choice question format. Each component serves a specific purpose:\\
\textbf{Question \{question\}} This section contains the main question extracted from the dataset.\\
\textbf{Choices (\{a\}, \{b\}, \{c\}, \{d\}):} These represent the options provided as answers for the question within the dataset. \\
\textbf{Correct Answer \{label\}) \{answer\}} This section indicates the correct label and its corresponding answer from the dataset. \\
This structured format enables the model to comprehend and process each question along with its associated choices and correct answer during the fine-tuning training process. 
During the \textit{inference phase}, the same prompt is reproduced, with the sole distinction of a blank space within the Answer section. This deliberate inclusion of a blank space aims to support the model's text generation process. In inference, the model is tasked with generating the correct answer using the information presented in the prompt. This setup enables the model to dynamically generate responses, utilizing its comprehension of the question and the contextual details provided within the prompt.

\section{Assessment and Insights on Dataset Quality}
\label{sec:quality}
Upon reviewing our incorrect predictions across both sub-tasks, subsequent to the task organizer releasing the labels for the test split of the datasets, we reached several conclusions. 
Across all triplets, encompassing original, semantic, and context reconstruction statements, we observe a considerable degree of ambiguity in various patterns. This ambiguity often leads to inconsistent selection of correct answers, even when answered by humans. This underscores the need for clearer formulation of questions and unambiguous expression to enhance the accuracy of model predictions. Another notable pattern we identified pertains to the quality control of semantic reconstruction in certain questions. In these instances, some words were not replaced with accurate synonyms, resulting in a shift in the definition of the brain teaser presented by the question. While this may not inherently be problematic, the dataset's correct answers remained unchanged compared to the original version of the question. This discrepancy suggests that the alteration in question definition went unnoticed by the task organizers, leading to some erroneous predictions by our model, when in reality the correct context of the provided multiple-choice statement was captured by our system. The two observations above highlight the inherent difficulty in generating clear and precise brain teasers, as well as the challenge that models face in understanding them. In the above scenarios, our top-performing model either detects the presence of a contradiction in the questions and opts to select "None of above," as elucidated in its brief and explanatory justification, or it provides an incorrect answer based on the dataset's answer but correctly reflects the problem context, which may have been altered due to inadvertent synonym usage.

\end{document}